# BINGO: A Novel Pruning Mechanism to Reduce the Size of Neural Networks

Aditya Panangat
Flower Mound High School
adipanangat@gmail.com

## Abstract

Over the past decade, the use of machine learning has increased exponentially. Models are far more complex than ever before, growing to gargantuan sizes and housing millions of weights. Unfortunately, the fact that large models have become the state of the art means that it often costs millions of dollars to train and operate them. These expenses not only hurt companies but also bar non-wealthy individuals from contributing to new developments and force consumers to pay greater prices for AI. Current methods used to prune models, such as iterative magnitude pruning, have shown great accuracy but require an iterative training sequence that is incredibly computationally and environmentally taxing. To solve this problem, BINGO is introduced. BINGO, during the training pass, studies specific subsets of a neural network one at a time to gauge how significant of a role each weight plays in contributing to a network's accuracy. By the time training is done, BINGO generates a significance score for each weight, allowing for insignificant weights to be pruned in one shot. BINGO provides an accuracy-preserving pruning technique that is less computationally intensive than current methods, allowing for a world where AI growth does not have to mean model growth, as well.

## Motivation

Machine learning developers like to go big, training models with millions of neurons. In fact, according to Howarth (2024), OpenAI's GPT-4 has 1.8 trillion parameters. Models are expected to be large to meet performance and accuracy standards; however, as these models grow, a few key problems arise.

First, training becomes incredibly expensive. Training billions of weights means performing forward passes to compute activations, backward passes to calculate gradients for all associated parameters, and updating these parameters iteratively across potentially millions of data samples, which is incredibly computationally expensive. According to Li (2020), it costs OpenAI about 4.6 million dollars to train their GPT-3 model just once. Moreover, according to Slowik (2023), because of the sheer size of Open AI's models, they can cost up to 1.5 million dollars per year for operational usage.

Second, this is a great barrier to accessibility. New developers are unable to afford these computational costs, preventing them from industry entry. The businesses that can afford this expense pass this cost down to consumers, increasing the price of machine learning models for clients. This is detrimental when it comes to buying machine learning models essential to crucial practices, such as illness detection or education.

Third, copious amounts of energy are used, hurting the environment via emissions. According to Hao (2019), training a model just once can emit more than 626,000 pounds of $CO_2$, with Patterson et. al (2021) finding that GPT-3 causes the emission of 502 metric tons of $CO_2$ per training session.

**Table A1**, created by Strubell *et al.* (2020), shows the energy efficiency of training models. CO2e estimates carbon dioxide emissions (lbs) produced by training a model once. Cloud computing cost resembles the cost (USD) of training a model once.

| Model | Hardware | Power (W) | Hours | kWh·PUE | $CO_2e$ | Cloud compute cost |
|---|---|---|---|---|---|---|
| $Transformer_{base}$ | P100x8 | 1415.78 | 12 | 27 | 26 | $41–$140 |
| $Transformer_{big}$ | P100x8 | 1515.43 | 84 | 201 | 192 | $289–$981 |
| ELMo | P100x3 | 517.66 | 336 | 275 | 262 | $433–$1472 |
| $BERT_{base}$ | V100x64 | 12,041.51 | 79 | 1507 | 1438 | $3751–$12,571 |
| $BERT_{base}$ | TPUv2x16 | — | 96 | — | — | $2074–$6912 |
| NAS | P100x8 | 1515.43 | 274,120 | 656,347 | 626,155 | $942,973–$3,201,722 |
| NAS | TPUv2x1 | — | 32,623 | — | — | $44,055–$146,848 |
| GPT-2 | TPUv3x32 | — | 168 | — | — | $12,902–$43,008 |

Frankle and Carbin (2018) at MIT CSAIL articulate a hypothesis dubbed *the lottery ticket hypothesis*: they substantiated that dense, randomly-initialized, feed-forward networks contain subnetworks (called winning tickets) that—when trained in isolation—reach test accuracy comparable to the original network in a similar number of iterations. Effectively, they argue that within every dense, trained neural network, there lies certain "winning tickets"–smaller networks that can achieve the same accuracy as the full model. Thus, solving the problems that arise from the existence of enormous models boils down to one question: how can these smaller models–these winning tickets–be found?

### Related Work

The most cost-effective approach that has been explored is called neural network pruning. Pruning relies on finding weights in a network that do not contribute significantly to the model's accuracy. In magnitude-based pruning, the most common approach, after training is complete, weights with values closest to zero are assumed to be insignificant and are removed. However, Rad and Seuffert (2024) show that this technique of one-shot pruning has been shown to drastically reduce accuracy at higher pruning rates. Even pruning just 10% of a network this way has been shown to result in accuracy dropping by 30 percentage points. Although pruning is efficient, it sacrifices accuracy.

A well-performing pruning method has been found by MIT researchers Frankle and Carbin (2018). They propose a nuanced version of *iterative magnitude pruning* (IMP). IMP is the best-performing pruning method so far established, and it functions like so:

A. A dense neural network is initialized, with each weight holding a randomly generated number as its value. These initial weights are stored. This network is then trained to completion.
B. Then, this 3-step process is performed iteratively:
   1. Weights closest to zero are pruned (removed) from the network, creating a smaller, pruned structure for the network
   2. Every weight in this new, pruned structure is reset to its initial assignment weight (saved in step A)
   3. This pruned structure is trained once again until it achieves accuracy similar to the original model

The process of setting the weights of the pruned architecture back to their initial weights and retraining allows for the unimportant weights to be identified repeatedly as the model is pruned slowly. On the MNIST dataset, Frankle and Carbin (2018) found that this version of IMP was able to remove 96% of the network's weights while maintaining the same test accuracy. However, the problem with this approach is that the neural network must be trained and retrained iteratively in order to prune. Training a model just once is costly (4.6 million dollars to train GPT-3). Hence, IMP is too expensive.

## Methodology

**Bingo Overview**

Although IMP works extraordinarily well, its major pitfall is the fact that it requires multiple training sessions in order to complete pruning. IMP, as a response to the lottery ticket hypothesis, acts in a way similar to the lottery: all the lottery numbers are printed, and only then is a winner found.

The ideal technique to reduce the size of models, then, is one that does not require additional training sessions to be done. So, I introduce BINGO, a novel neural network pruning technique that addresses the lottery ticket hypothesis not by iteratively finding unnecessary weights through additional training sessions, but by identifying which weights will be unnecessary while the original model itself is training. Rather than finding winning tickets after all the winning numbers are called out, people find BINGO one letter at a time while the game is taking place. Simply put, BINGO solves the problem of needing additional iterative training sessions for model pruning by collecting information during the original training session. With this information, after the model is trained, all unimportant weights shall be pruned from the model in one shot. This means that computationally expensive, repetitive training and resetting of the model are not necessary, drastically reducing the computational and environmental cost required to run fast, efficient models. When models are faster for cheaper, machine learning is a less environmentally taxing and more accessible field.

**Bingo Design**

BINGO achieves computationally efficient pruning by contributing 2 novel, creative, and effective mechanisms to the literature.

1. A technique to isolate subsets of a neural network called *lottery ticket searching*.

This technique borrows from neural network dropout, a regularization technique proposed by Srivastava *et al.* (2014), which attempts to make models more generalizable by deactivating a random subset of weights during each training pass. In a similar way, lottery ticket searching, just after each training pass, sets a subset of *weights* back to their initial values. It is important to note that this process does not truly affect the weights in the network. Instead, just after each real training pass, this technique simply simulates a "fake training pass," setting some weights back to their initial assignments to simply keep score of what *would have* happened if those were the true weights.

2. A new measurement called *weight significance score*.

Based on the way that the accuracy of a model is calculated to change during each lottery ticket search, BINGO calculates a weight significance metric for each weight in the neural network. The weight significance for each weight is scaled from 0 to 1, with higher scores representing weights that are believed to contribute more significantly to model accuracy. This metric is calculated using data that is gathered from lottery ticket searching, which keeps track of what happens to accuracy when certain subsets of weights are set back to their initial, untrained values. Data from each lottery ticket search helps calculate a total significance score for every weight that was temporarily reset to its initial value. Initially, weight significance is set to 0.5 for every weight. After a lottery ticket search, the following calculation for weight significance ($S_w$) is used to update significance scores for the weights that were temporarily reset to their initial values:

$$S_w = \frac{S_p + \left(1 - \frac{|A_{before} - A_{after}|}{A_{before}}\right)}{2}$$

Where:
- $S_w$ is the weight's new weight significance score
- $S_p$ is the weight's previous weight significance score
- $A_{before}$ is the model's accuracy before resetting the weight to its initial value
- $A_{after}$ is the model's accuracy after resetting the weight to its initial value

Finally, after the model is finished training, BINGO will prune weights until a certain minimum model accuracy is reached, pruning weights of lowest significance score first.

**BINGO**

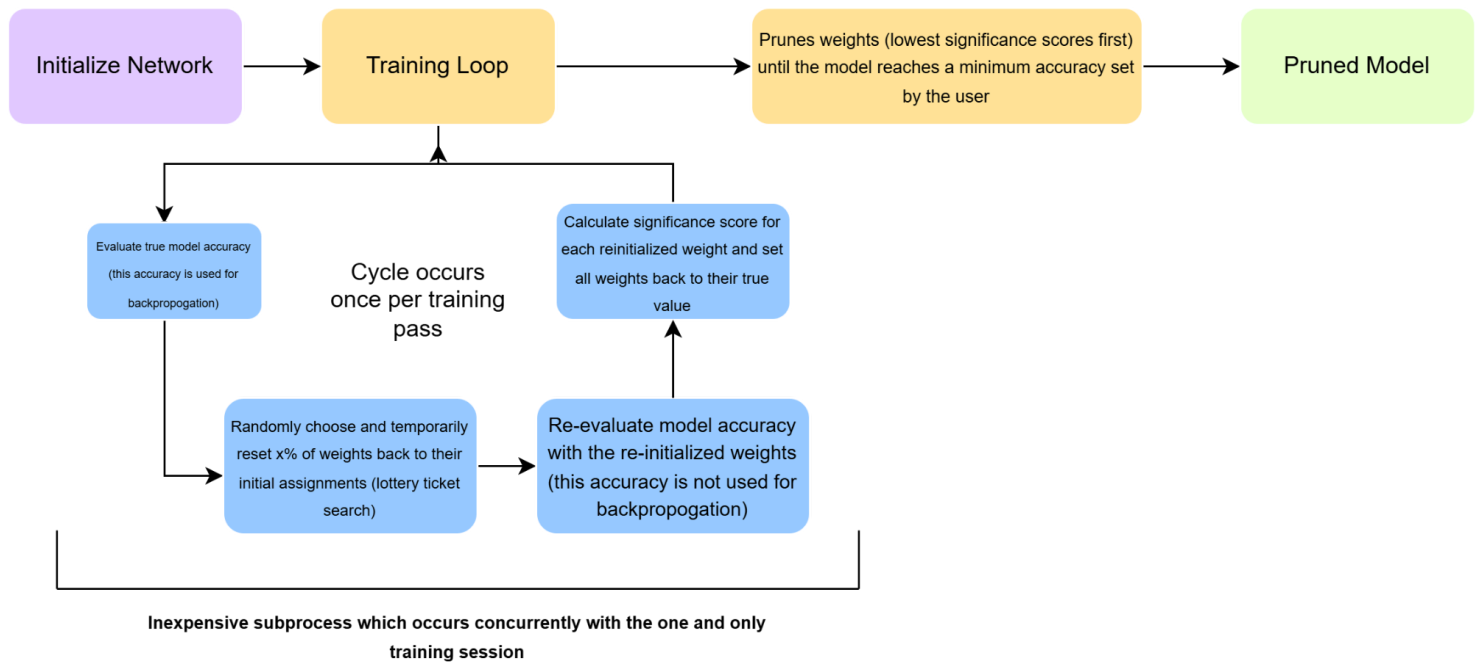

## Results

An ML model was trained on the MNIST dataset produced by LeCun (2009). The following results table compares pruning of this model by IMP and BINGO.

| Method | Training + Pruning Time (min) | % of Total Weights Pruned |
|---|---|---|
| IMP | 122 | 85% |
| BINGO | 17 | 71% |

Table 1: Pruning results on an MNIST-trained model. IMP and BINGO were both programmed to prune until the model fell to 90% accuracy from an initial 97%

**Discussion and Conclusion**

       When told to prune a model down to the same accuracy, I find that BINGO prunes 86.07% faster than IMP does. At the same time, IMP was able to prune 14 percentage points more weights than BINGO was able to. Compared to IMP, BINGO serves as a mechanism to prune weights that is similar in efficacy and far less computationally taxing.

       Because BINGO only takes 17 minutes to prune a model that IMP would take 122 minutes to prune, it serves as a superior method to prune down machine learning models. Ultimately, this makes ML as a field less environmentally taxing and drives down the computational cost required to run ML and AI models.